\documentclass[conference]{IEEEtran}
\IEEEoverridecommandlockouts
\usepackage{cite}
\usepackage{amsmath,amssymb,amsfonts}
\usepackage{algpseudocode} 
\usepackage{algorithm}
\usepackage{graphicx}
\usepackage{textcomp}
\usepackage{xcolor}
\usepackage{caption}
\usepackage{subcaption}
\def\BibTeX{{\rm B\kern-.05em{\sc i\kern-.025em b}\kern-.08em
    T\kern-.1667em\lower.7ex\hbox{E}\kern-.125emX}}
\begin{document}

\title{Practical Layout-Aware Analog/Mixed-Signal Design Automation with Bayesian Neural Networks 
}

\author{\IEEEauthorblockN{Ahmet F. Budak}
\IEEEauthorblockA{\textit{The University of Texas at Austin} \\
Austin, TX, USA \\
ahmetfarukbudak@utexas.edu}
\and
\IEEEauthorblockN{Keren Zhu}
\IEEEauthorblockA{$\textit{The University of Texas at Austin} $\\
Austin, TX, USA \\
keren.zhu@utexas.edu}
\and
\IEEEauthorblockN{ David Z. Pan}
\IEEEauthorblockA{\textit{The University of Texas at Austin} \\
Austin, TX, USA \\
dpan@ece.utexas.edu}

}

\maketitle

\begin{abstract}
The high simulation cost has been a bottleneck of practical analog/mixed-signal design automation. Many learning-based algorithms require thousands of simulated data points, which is impractical for expensive to simulate circuits. We propose a learning-based algorithm that can be trained using a small amount of data and, therefore, scalable to tasks with expensive simulations. Our efficient algorithm solves the post-layout performance optimization problem where simulations are known to be expensive. Our comprehensive study also solves the schematic-level sizing problem. For efficient optimization, we utilize Bayesian Neural Networks as a regression model to approximate circuit performance. For layout-aware optimization, we handle the problem as a multi-fidelity optimization problem and improve efficiency by exploiting the correlations from cheaper evaluations. We present three test cases to demonstrate the efficiency of our algorithms. Our tests prove that the proposed approach is more efficient than conventional baselines and state-of-the-art algorithms.
\end{abstract}


\section{Introduction}
\begin{figure}[t]
\centering
\includegraphics[scale=0.4]{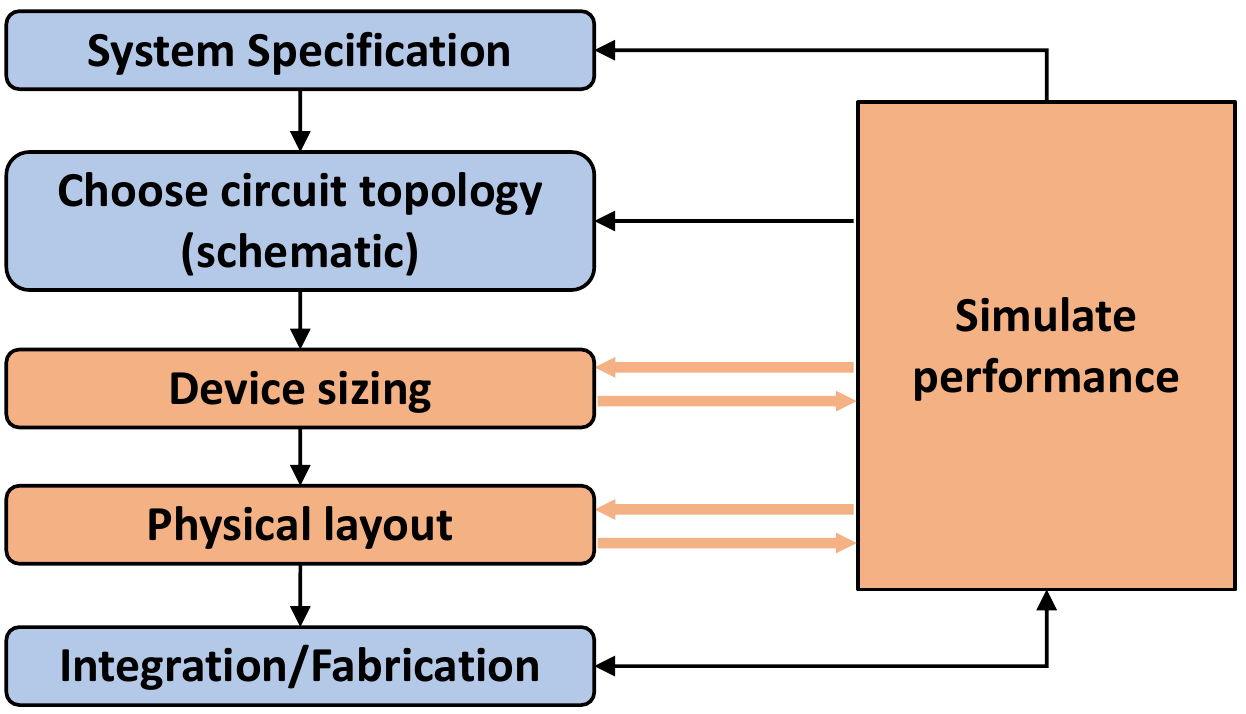}
\vspace{-1mm}
\caption{AMS Design Flow}
\label{fig:analog_design_flow}
\vspace{-6mm}
\end{figure}

Analog/Mixed-signal~(AMS) integrated circuit~(IC) design typically follows a process flow visualized in Figure~\ref{fig:analog_design_flow}. A combination of designer experience and computer simulation feedback is iterated to determine the design that meets the performance requirements. A large portion of design time is spent on the sizing and layout phases, where multiple iterations are possible due to potential loop-backs in the design flow. This is a labor-intensive process in industry practice with little to no automation. To address this costly exercise, a considerable effort in academia is focused on introducing automated solutions.

Analog sizing automation is the task of optimizing AMS design variables, e.g., transistor widths, lengths, resistor, and capacitor values. The aim is to satisfy the performance constraints and optimize the design objective. In general, sizing automation is run through schematic-level simulations. However, AMS IC performance is also sensitive to layout implementation~\cite{10.1145/3569052.3578929}. Especially in the advanced process nodes, layout-induced parasitics may greatly affect the final design performance. Therefore, sizing the AMS design variables considering the layout effects is also crucial.

The majority of the recent sizing and post-layout performance optimization algorithms have simulation feedback in the loop. Due to advanced scaling, simulations are required to obtain accurate performance evaluations. Simulation-based AMS automation algorithms adapted various methods from the optimization and Machine Learning~(ML) communities. The earlier approaches include population-based methods such as particle swarm optimization~\cite{Vural2012AnalogCS} and evolutionary algorithms~\cite{Liu:2013:ADA:2526263}. Although these algorithms have good convergence behavior, they are inefficient in sampling since they explore the design space randomly. To mitigate sample inefficiency, model-based methods gained popularity~\cite{GASPAD,Lyu:2018:MBO:3195970.3196078, 9775179}. These methods employ surrogate-models between the solution space and performance space and provide efficiency in exploring the solution space. A typical surrogate model is Gaussian Process Regression (GPR)~\cite{Rasmussen:2005:GPM:1162254}, which is a well-studied model in Bayesian Optimization (BO) field~\cite{NIPS2012_05311655} and is adapted by several analog sizing algorithms. The main drawback of GPR modeling is its computational complexity.

Recent research trend in analog sizing introduces ML to simulation-based methodology~\cite{Analog_book2022_Budak}. However, the literature review reveals that most of these methods require thousands of simulation data to train Deep Neural Network~(DNN) models that approximate the relations between the design variables and the performance metrics. Therefore, the practicality of these algorithms is severely reduced when the optimization task has a high simulation cost. For example, drawing/generating the layout, extracting the parasitics, and running post-layout simulations is typically an expensive procedure. Therefore, optimization algorithms designed for schematic-level sizing can not be adapted by simply changing how data is generated.

This paper presents a Machine Learning-based simulation-in-the-loop automation method for the AMS design problem. Overall, we formalize two stand-alone recipes for schematic-level sizing and post-layout performance optimization, i.e., layout-aware sizing. We integrate the state-of-the-art analog layout generator, MAGICAL~\cite{MAGICAL_ICCAD19_Xu}, into our flow to handle layout-aware sizing. Our algorithms do not assume the pre-existence of any dataset, and we generate all training data during the optimization. We employ Bayesian Neural Networks~(BNN) for modeling design performances. Bayesian Neural Networks allow error quantification, and compared to Deep Neural Networks, BNN are shown to be effective in handling scarce datasets and preventing overfitting~\cite{Goan_2020}. Therefore, BNN can be trained on smaller datasets, significantly improving the practicality and scalability. We also introduce a batch-optimization framework and design space sampling strategy that is compatible with BNN modeling. Further, when optimizing the design variables based on post-layout performance, we exploit the correlation between schematic-level simulations and post-layout simulations. Our algorithm introduces a co-learning scheme that reduces the need for costly post-layout simulations and boosts efficiency even further. We compile our contributions as follows:
\begin{itemize}
    \item We use Bayesian Neural Network-based modeling to obtain performance approximations. Different learning strategies are adapted for schematic-level sizing and post-layout performance optimization.
    \item We adopt a scalable sampling strategy and query the optimization batches by utilizing a trust region and Thompson sampling. 
    \item The post-layout sizing is handled as a multi-fidelity optimization problem, and an efficient co-learning strategy is developed.
    \item The efficiency of the proposed methods is demonstrated on three circuits by providing comparisons to previous state-of-the-art.
\end{itemize}

The rest of the paper is organized as follows.
Section~\ref{sec:background} introduces the backgrounds and previous work.
Section~\ref{sec:algo} describes our algorithms for handling schematic-level sizing and post-layout performance-based sizing problems.
Section~\ref{sec:results} provides the experiments on circuit examples to demonstrate the efficiency of our algorithms.
Finally, Section~\ref{sec:conclusion} concludes the paper.
\section{Background \& Related Work} \label{sec:background}

\begin{figure*}[t]
\centering
\includegraphics[scale=0.69]{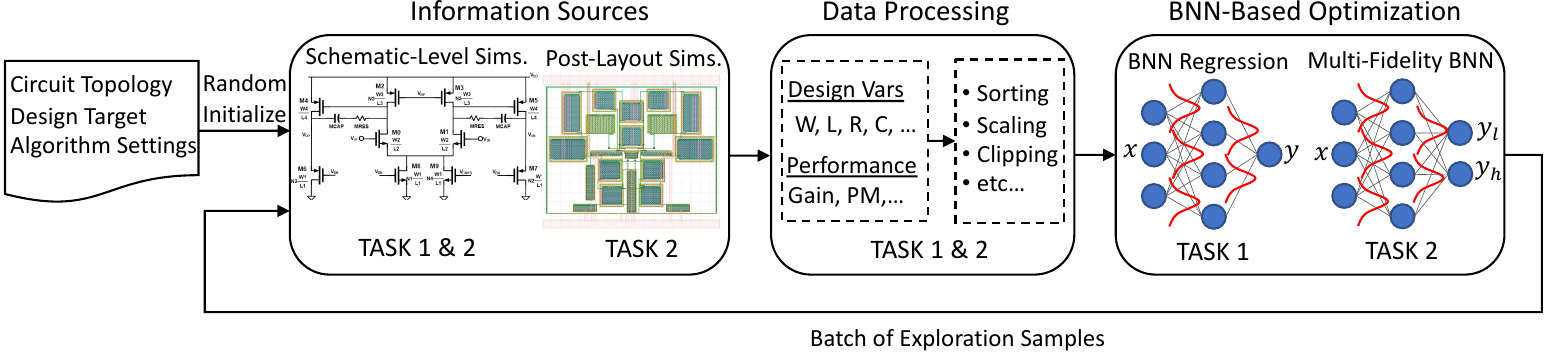}
\caption{Proposed AMS Automation Framework}
\vspace{-4mm}
\label{fig:overall_framework}
\end{figure*}

In this section, we first formally define the AMS design automation problem. Then we review the recent approaches to schematic-level sizing and layout-aware sizing. We summarize the state-of-the-art algorithms' advantages and shortcomings.

\subsection{Problem Formulation}
\label{sec:background:problem}
In this paper, we assume that the existence of post-layout performance implies the existence of schematic-level performance values. However, the reverse implication does not hold. We formulate the AMS schematic-level sizing and layout-aware sizing task as a constrained optimization problem succinctly as below.
\vspace*{-1mm}
\begin{equation}
    \vspace*{-1mm}
    \label{eq:prob_formulation}
    \begin{aligned} \operatorname{minimize}\text{ } & f_{0}(\mathbf{x}) \\ \text { subject to } & f_{i}(\mathbf{x}) \leq 0 \quad \text { for } i=1, \ldots, m 
    \end{aligned}
\vspace*{-1mm}
\end{equation}
where, $\mathbf{x}\in\mathbb{R}^{d}$ is the parameter vector and $d$ is the number of design variables of sizing task. Thus,  $\mathbb{R}^{d}$ is the design space. $f_0(\mathbf{x})$ is the objective performance metric we aim to minimize. Without loss of generality, we denote $i^\text{th}$ constraint by $f_i(\mathbf{x})$. Notice that if the problem is schematic-level optimization, the $f_i$ values are obtained from schematic simulations. If the problem is post-layout optimization, the $f_i$ values are determined by post-layout simulations.
	
Through this paper, we will evaluate the quality of a design by defining a Figure of Merit (FoM) in the following form:
\vspace*{-2mm}
\begin{equation}
    \vspace*{-1mm}
    \label{eq:scalarizationfunc}
    \text{FoM}(\mathbf{x}) = w_0\times f_0(\mathbf{x}) + \sum_{i=1}^{{m}} \mathrm{min}\left(1, \mathrm{max}(0, w_i \times f_i(\mathbf{x}))\right)
    \vspace*{-1mm}
\end{equation}
where $w_i$ is the weighting factor. Note, a $\mathrm{max(\cdot)}$ clipping is used for equating designs after constraints are met, and $\mathrm{min(\cdot)}$ is used to prevent single constraint violation from dominating $\text{FoM}$ value.
	
\subsection{Schematic-Level Sizing} \label{sec:background:sizing}
The recent methods for AMS sizing can be collected under two algorithm classes: Bayesian Optimization methods and Deep Learning methods.

Bayesian Optimization methods are tested on AMS problems and are proven to be sample efficient. For example, GASPAD~\cite{GASPAD} is a hybrid algorithm using a combination of evolutionary space exploration and GPR surrogate-based selection. WEIBO~\cite{Lyu:2018:MBO:3195970.3196078} method also employs GPR as a surrogate and introduces a Bayesian Optimization framework where a weighted acquisition function is tailored to comply with the performance-constrained nature of sizing problem. In~\cite{9775179}, the authors introduced a multi-fidelity GPR algorithm where the fidelity of the performance is varied with the simulation accuracy. However, this work did not address the layout effects. The disadvantage shared by all GPR models is their cubic complexity to the number of samples, $\mathcal{O}(N^3)$.

Deep Learning based sizing methods includes supervised learning and reinforcement learning~(RL) methods~\cite{9434374,Wang2020GCNRLCD,Settaluri2020AutoCktDR,budak2021dnnopt, 9643445}. GCN-RL~\cite{Wang2020GCNRLCD} is a Graph Neural Network algorithm where state representation is built via device index, type, and selected electrical properties. They also propose methods to transfer the optimization experience between different topologies and processes. However, their training graphs show that they use up to $10^4$ simulations for sizing academic circuits. AutoCkt~\cite{Settaluri2020AutoCktDR} is a discrete action space policy gradient method. The RL agent is trained on different optimization tasks where the task is randomly sampled from a predefined set. The trained agent is then tested for the particular tests during deployment. We also observe from the training graphs that AutoCkt requires up to $10^5$ simulated samples for training. In~\cite{8942174}, the authors successfully applied BNN on multi-objective analog sizing. However, they did not consider handling constraints, and layout effects are ignored. DNN-Opt~\cite{budak2021dnnopt} is introduced as an RL-inspired supervised learning optimization method that shows high sample efficiency and can be trained during optimization. It uses less than a thousand iterations to optimize academic benchmarks. DNN-Opt does not quantify the variance on approximated values and has no methodic way to balance exploration/exploitation during design space exploration.

\subsection{Post-Layout Based Sizing} \label{sec:background:layout}
Several works in AMS sizing proposed solutions to include layout-induced parasitics. The studies proposed in~\cite{LOURENCO2016316} and~\cite{Analog_ICCAD19_Hakhamaneshi} embedded a layout generator in the automation loop, and performance metrics to be optimized are obtained through post-layout simulations. However, they did not consider the correlations between the schematic-level and post-layout simulations; therefore, their efficiency is limited. The work in~\cite{Analog_DATE21_Liu} employs a less accurate parasitic prediction during sizing, so the finalized post-layout performance is not guaranteed. In~\cite {9643445}, the authors propose a Transfer Learning strategy where a DNN is first trained on schematic-level simulations. Then this knowledge is transferred to improve the learning of a relatively small number of post-layout data. Although this work provides a suggestion to improve the efficiency of post-layout optimization, it requires up to $5\times10^3$ schematic-level data for initial DNN training, which suffers from the scalability concerns mentioned before. In summary, a scalable solution to optimize AMS design parameters under layout parasitics is yet to be studied.
\section{Analog/Mixed-Signal IC Automation Flow}
\label{sec:algo}
In this work, we provide solutions to two problems in Analog/Mixed-Signal design automation: schematic-level sizing automation~(Task 1) and layout-aware sizing automation~(Task 2). The high-level frameworks of proposed solutions to both tasks are summarized together in Figure~\ref{fig:overall_framework}. Section III-A introduces and elaborates on the core principles that complete our proposed automation flow for schematic-level sizing tasks. Then, in section III-B, we explain how to solve the post-layout performance optimization problem efficiently by transforming the BNN learning scheme.

\subsection{Schematic-Level Sizing Automation}
We propose a BNN-based sizing algorithm to optimize AMS design on schematic-level simulations. The complete flow of the proposed approach is summarized in ~Algorithm~\ref{alg:bnn_sizing}. The algorithm starts with sampling random points in the design space and simulating them via the SPICE simulator. An initial dataset for training the BNN performance model is built from these samples. Then a trust-region state is initialized before algorithm iterations start. The trust region determines the bounds of the exploration space. The following subsections will provide more details regarding the BNN modeling and trust-region search.

Our algorithm models each performance metric at each optimization iteration with an individual BNN model. Then a batch of samples is collected based on the posterior realization of points lying inside the trust region. Candidate design performance realizations are obtained using the Thompson sampling method, and the candidates are ranked based on the utility values~(FoM). A batch of $q$ points is collected, and their real performances are obtained through simulation. The new data is added to the database, and the trust region is updated based on the real simulation outputs of the last batch.

\begin{algorithm}[h]
\caption{BNN-Based Sizing Algorithm}
\label{alg:bnn_sizing}
\begin{algorithmic}[1]
  \Require An initial sample set $\mathbf{X}^\mathrm{init}$ of $N_\mathrm{init}$ designs and their evaluations $f(\mathbf{X}^\mathrm{init})$
  \State Assign the solution with maximum utility
  \State Initialize trust-region state
  \For {$t = 1, 2, \dots,t_{max}$}
    \State Train BNN for each performance metric
    \State Generate $r$ candidate points $x_1, \dots , x_r \in\Omega$
in the trust region.
    \For {$b = 1, 2, \dots,q$}
    \State For each of the $r$ points of the next batch, sample a realization
$\{(\hat{f}\left(x_{i}\right), \hat{f}_{1}\left(x_{i}\right), \ldots, \hat{f}_{m}\left(x_{i}\right))^{T} \mid 1 \leq i \leq r\}$ from the posterior over each candidate and add a point of maximum utility to the batch.
    \EndFor
    \State Simulate the $q$ new query points and obtain specs $f(\mathbf{X}_t)$ via SPICE sims
    \State Update database with new designs and evaluations
    \State Update trust region state by comparing the current best and $f(\mathbf{X}_t)$
  \EndFor
   \State \Return The design with the highest utility
\end{algorithmic}
\vspace*{0mm}
\end{algorithm}

\subsubsection{Performance Modeling with Bayesian Neural Networks}
We base our Bayesian Neural Network regression method on the assumption that the observed function values follow a Gaussian distribution and the probabilistic model on the observations are in the following form:
\begin{equation}
    p(f(x)\mid x,\theta) = \mathcal{N}\left(\phi(x;\theta),\tau^{-1}\right)
\end{equation}
where $\theta$ is the BNN parameters, i.e., weights and biases, $\phi(x;\theta)$ is the output of the BNN with parameters $\theta$ and $\tau$ is the noise parameter.
We assign a standard Gaussian prior distribution over each element of the NN parameters, $\mathcal{\theta}$, and a Gamma prior over each noise precision, $p\left(\tau\right)=\operatorname{Gam}\left(\tau \mid a_{0}, b_{0}\right)$. Let define $y_n = f(x_n)$. Given
the dataset $\mathcal{D}=\left\{\left(\mathbf{x}_{n}, y_{n}\right)\right\}_{n=1}^N$, the joint probability of our model is given by
\begin{equation}
\resizebox{.999\hsize}{!}{
    $p(\mathcal{\theta}, \mathcal{Y}, \tau, \mid \mathcal{X})=\mathcal{N}(\operatorname{vec}(\mathcal{\theta}) \mid \mathbf{0}, \mathbf{I}) p\left(\tau\right) \prod_{n=1}^N \mathcal{N}\left(y_{n} \mid \phi\left(\mathbf{x}_{n}\right), \tau^{-1}\right)$}
\end{equation}
where $\mathcal{X}=\left\{\mathbf{x}_{n}\right\}, \mathcal{Y}=\left\{y_{n}\right\},\text{ and vec}(\cdot)$ is vectorization. 

Due to its unbiased, high-quality uncertainty quantification, we use Hamiltonian Monte Carlo (HMC)~\cite{1206.1901} sampling to perform posterior inference and generate samples of $\theta^i\sim p(\theta\mid\mathcal{D})$ from the posterior of BNN parameters. Then, using the samples of $\theta$, we make a Gaussian approximation to the function value as follows:
\begin{align}
\label{Eq:predictive_model}
\begin{split}
    \mu(f(x)\mid\mathcal{D})&=\frac{1}{M}\sum_{i=1}^M\phi(x;\theta^i)\\
    \sigma^2(f(x)\mid\mathcal{D})&=\frac{1}{M}\sum_{i=1}^M\left(\phi(x;\theta^i) - \mu(f(x)\mid\mathcal{D})\right)^2 + \frac{1}{\tau}
\end{split}
\end{align}

where $\mu$ is the mean and $\sigma^2$ is the variance approximation. 

\subsubsection{Trust-Region Search Engine}
We follow the trust region approach introduced in~\cite{ErikssonP21} and confine the candidate points locally. The trust-region assigns a localized subset of the search space and proceeds in rounds. We denote the trust region by $\Omega$. In each round, a batch of $q$ designs in $\Omega$ are selected by the BNN algorithm and then simulated in parallel. Note that this procedure is easily extended
to asynchronous batch evaluations, and we adapt asynchronous evaluation for the multi-fidelity BNN algorithm (will be discussed), where evaluation times show significant differences. The trust-region is centered around the best design explored, i.e., the design with minimum FoM where the ties are handled according to the design objective. This approach mitigates common issues of Bayesian optimization in high-dimensional settings, where popular acquisition functions fail to focus on promising regions and spread out samples due to large prediction uncertainty.

\textbf{Thompson Sampling-based Exploration}: We employ Thompson sampling to obtain performance approximations for untested design candidates. Thompson sampling scales to large batches at low computational cost and has shown to be as effective as the expected improvement acquisition function\cite{ErikssonP21}. Further, the Thompson sampling naturally extends to constrained settings which is usually the case for AMS automation. To select a point for the next batch, we sample $r$ candidate points in $\Omega$. Let $x_1, \dots, x_r$ be the sampled candidate points. Then we use the predictive model given in Equation~\ref{Eq:predictive_model}, and sample a realization $(
\hat{f}_0(x_i), \hat{f}_1(x_i), . . . , \hat{f}_m(x_i))^T$
for all $x_i$ with $1 \leq i \leq r$ from the respective posterior distributions on the functions $f_0, f_1,\dots, f_m$. Let $\hat{F}=\left\{x_{i} \mid \hat{f}_{j}\left(x_{i}\right) \leq 0\right. \text{for}\left.1 \leq j \leq m\right\}$ be the set of points
whose realizations are feasible. If \(\hat{F} \neq \emptyset\) holds, we
select an $\text{argmin}_{x \in \hat{F}} \hat{f}(x)$, i.e., the design with minimum objective. Otherwise, we select a
point of minimum total violation based on the FoM definition given in Equation~\ref{eq:scalarizationfunc}.

\textbf{Maintaining the trust-region:} We initialize a trust region as a hypercube with side length L around the maximum utility point. As the optimization progresses, we track the number of successes $n_s$ and failures $n_f$ since the last time the trust-region is updated. A success is when the algorithm improves the solution quality, and by construction, this point must be inside the trust region. We call it a failure when the last batch of simulated designs is worse than the current best solution. The center C of the trust region is updated as follows. If there exist feasible designs, the one with the minimum objective is assigned as the center. Otherwise, the design with minimum FoM, i.e., minimum scaled constraint violation, is chosen as the center. Therefore, the center of the trust-region is updated every time the design performance is improved. The side length of the trust region is updated as follows: if $n_s = \tau_s$
then the side length is set to $L = \min\{2L, L_{\mathrm{max}}\}$ and
we reset $ns = 0$. If $n_f = \tau_f$ , then we set $L = L/2$ and
$n_f = 0$. If the side length drops below a set threshold $L_{\mathrm{min}}$, we initialize a new trust region.

\subsection{Post-Layout Performance Optimization}
\begin{figure}[h]
\centering
\includegraphics[scale=0.4]{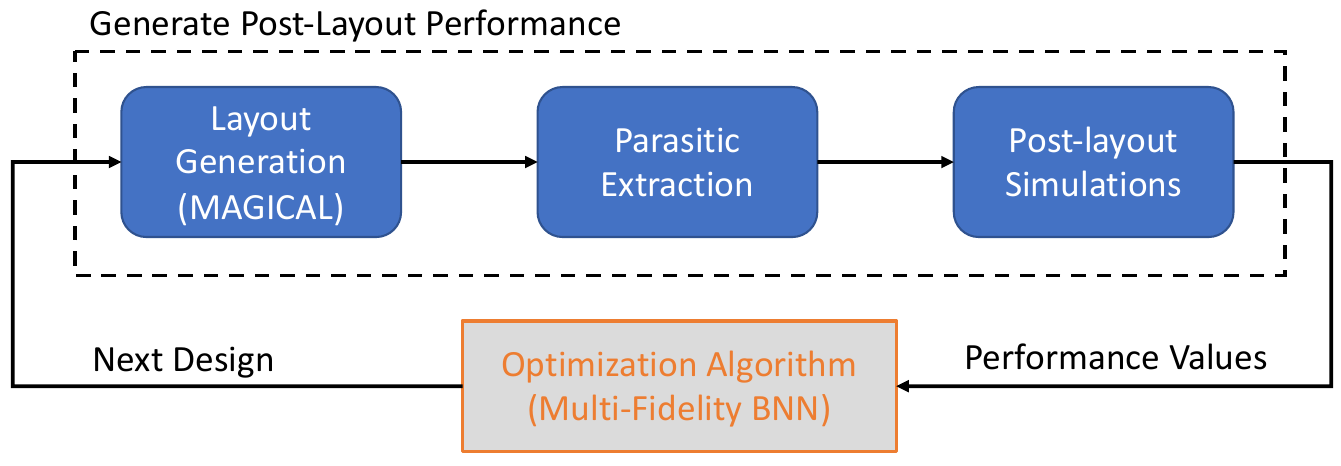}
\caption{Post-Layout Performance Based Optimization}
\label{fig:layout_in_flow}
\end{figure}

We start our discussion by defining the modifications necessary to automate post-layout performance-based AMS sizing. We tailor the classical sizing flow to include the post-layout effects on the performance during sizing. Instead of optimizing the design variables based on the schematic level simulations, we utilize the layout automation tool MAGICAL to modify performance evaluation steps. The suggested flow is shown in Fig~\ref{fig:layout_in_flow}. First, an automated layout is generated via MAGICAL to obtain the post-layout performance of each new design. This step is followed by parasitic extraction, and circuit simulations are run on the updated netlist with parasitic elements.

However, this new flow is much more expensive than the schematic-level sizing task since the additional steps (layout generation, parasitic extraction, and post-layout simulations) are typically computationally expensive. Therefore, methods are sought to further increase the efficiency of the (BNN-based) optimization algorithm. As a solution, we treat this problem as a multi-fidelity problem where we have access to two different information sources for calculating circuit performance metrics. Considering that the schematic-level simulations are less accurate approximations of post-layout level simulations, we define these information sources as schematic-level simulations having the lower fidelity and post-layout simulations are the highest fidelity.

We modify the BNN architecture to capture two levels of fidelities (Figure~\ref{fig:overall_framework}) at the output and propose a co-learning scheme similar to multi-task BNN learning~\cite{NIPS2016_a96d3afe}. The multi-fidelity BNN model has two output nodes where $\phi(x)[1]$ models the lower fidelity prediction, i.e., schematic-level performance prediction, and $\phi(x)[2]$ models the high fidelity prediction, i.e., post-layout performance prediction. Under the assumption that we have access to two levels of information sources, we denote the new dataset by $\mathcal{D}=\mathcal{D}_1\cup\mathcal{D}_2$, where $\mathcal{D}_i=\left\{\left(\mathbf{x}_{n}^k, y_{n}^k\right)\right\}_{n=1}^{N_k}$ and the joint probability of the updated BNN model is given by:
\begin{equation}
\resizebox{.999\hsize}{!}{
    $p(\mathcal{\theta}, \mathcal{Y}, \tau, \mid \mathcal{X})=p(\theta) p\left(\tau\right) \prod_{k=1}^2\prod_{n=1}^{N _k}\mathcal{N}\left(y_{n}^k \mid \phi\left(\mathbf{x}_{n}^k\right)[k], \tau^{-1}\right)$}
\end{equation}
where $p(\theta)=\mathcal{N}(\operatorname{vec}(\mathcal{\theta}) \mid \mathbf{0}, \mathbf{I})$ and $N_k$ is the number training points in given fidelity level $k$. The joint probability expression is a combination of the data sourced from both types of simulations; therefore, we utilize the full dataset to train multi-fidelity BNN. In this way, both fidelities are learned together, and the correlations between them are captured due to shared BNN parameters.

To handle the multi-fidelity problem, we adopt the following modifications to Algorithm 1:
\\
1) We train multi-fidelity BNN models using the whole history of simulations, $\mathcal{D}$.
\\
2) The trust-region centering and length updates are based on the post-layout simulation results, i.e., highest-level fidelity results.
\\
3) We determine the candidate selection by modifying the work of~\cite{NIPS2016_605ff764} where they propose an upper-confidence-bound selection criteria for a single objective BO. We obtain Thompson sampling-based realizations for each fidelity, i.e., $\{\hat{f}_i^{(1)}(\mathrm{x}),~\hat{f}_i^{(2)}(\mathrm{x}), ~\text{for }i=0,1,\dots,m\}$ where $\{1,2\}$ indicate the fidelity level (schematic-level simulations and post-layout level simulations) and then calculate the low fidelity and high fidelity FoM approximations, $FoM(\hat{f}^L(x))$ and $FoM(\hat{f}^H(x))$ using the corresponding realizations. The candidate selection is queried according to the following utility expression:
$$\mathcal{U}(x)= \max(FoM(\hat{f}^L(x))-\Delta,~FoM(\hat{f}^H(x)))$$
where $\Delta$ is the FoM difference between the samples with the best utility at each fidelity. In this step, we take a practical approach to convert two fidelities to each other by defining a reduction term and assign the conservative prediction as the utility value. Finally, the argmin selection is conducted on the candidate utility values to determine the next batch.
\\
4) The current literature on multi-fidelity Bayesian optimization lacks in handling large number of constraints. Therefore, we randomly assign the fidelity (simulation type) for selected candidates and leave the fidelity selection as future work. Note that this action does not prevent us from studying the benefits of multi-fidelity handling of layout-aware sizing. However, we sacrifice potential cost-aware improvements through intelligent fidelity selection.
\\

\vspace{-3mm}
\section{Experiments}
\begin{figure*}[t]
\centering
\includegraphics[scale=0.44]{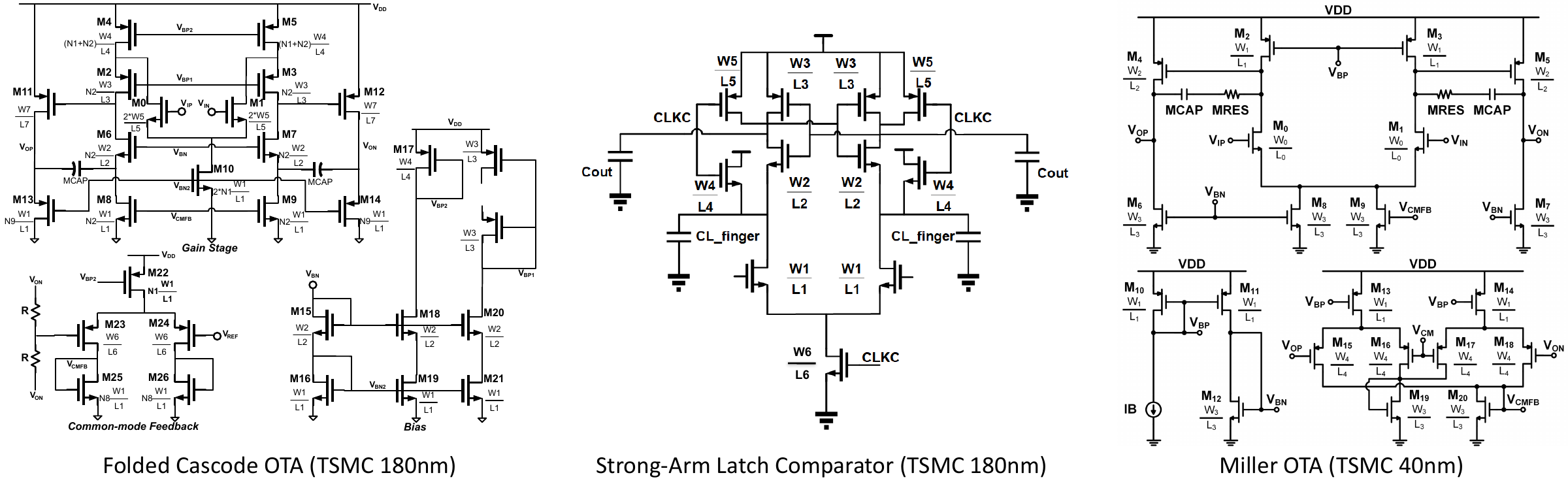}
\caption{Schematics of Tested AMS Circuits}
\label{fig:all_exp_cirs}
\end{figure*}

\textbf{Experiment Setup and Algorithm Settings:}
\label{sec:results}
We run our tests using 3 different AMS circuits designed with different technologies. A Two-Stage Folded Cascode Operational Transconductance Amplifier~(OTA), and a Strong-Arm Latch Comparator are designed with TSMC 180nm process and used to test schematic-level sizing algorithms. Then, we demonstrate the results for layout-aware algorithms on a Two-Stage Miller OTA. This circuit is designed in TSMC 40nm technology since the layout generator used in this work, MAGICAL, is crafted for TSMC 40nm. The schematic designs for these circuits are included in Figure~\ref{fig:all_exp_cirs}.

We run experiments to study the effectiveness of both of the proposed algorithms. First, we test for the schematic-level sizing algorithm, which is given by Algorithm 1, and we refer to our Bayesian Neural Network Based Bayesian Optimization algorithm as "BNN-BO". Then, we run tests for our post-layout performance-based sizing algorithm. Since we utilize a multi-fidelity BNN for this task, we will refer to this algorithm as "MF-BNN-BO".

We implemented several state-of-the-art baseline algorithms to compare and quantify the quality of our proposed algorithms. We selected the baseline algorithms to cover the different categories of approaches. We list the compared baseline algorithms as follows: 1) A differential evolution global optimization algorithm~(DE), 2) Bayesian Optimization with weighted expected improvement (BO)~\cite{Lyu:2018:MBO:3195970.3196078}, and, 3) RL-based sizing algorithm, DNN-Opt~\cite{budak2021dnnopt}. All algorithms are implemented using Python. We implemented DNN-Opt via PyTorch~\cite{NEURIPS2019_9015}, Bayesian Optimization algorithm is implemented using BoTorch~\cite{balandat2020botorch} package and BNN-BO and MF-BNN-BO are implemented using PyTorch and Hamiltorch~\cite{cobb2020scaling} packages.

We configured BNN-BO and MF-BNN-BO to evaluate a batch of $q=8$ designs in parallel. For fairness, DNN-Opt and BO are also configured to do parallel evaluations. Both our algorithms use 200 HMC samples to train BNN models. All BNN models are feedforward neural networks with 2 hidden layers and 100 nodes at each hidden layer. Trust-region is initiated with $L=0.8$ and $L_{min}$ and $L_{max}$ are chosen to be $0.5^4$ and $1.6$, respectively. Failure and success tolerances as chosen as $n_f=2$ and $n_s=3$. All experiments are run on the same machine using CPU for training learning~(DNN and BNN) models. During experiments, the model-based algorithms BO, DNN-Opt, and BNN-BO are run until exploring 500 designs, and DE is run for 5000 new samples.

\begin{figure}[h]
\centering
\includegraphics[scale=0.42]{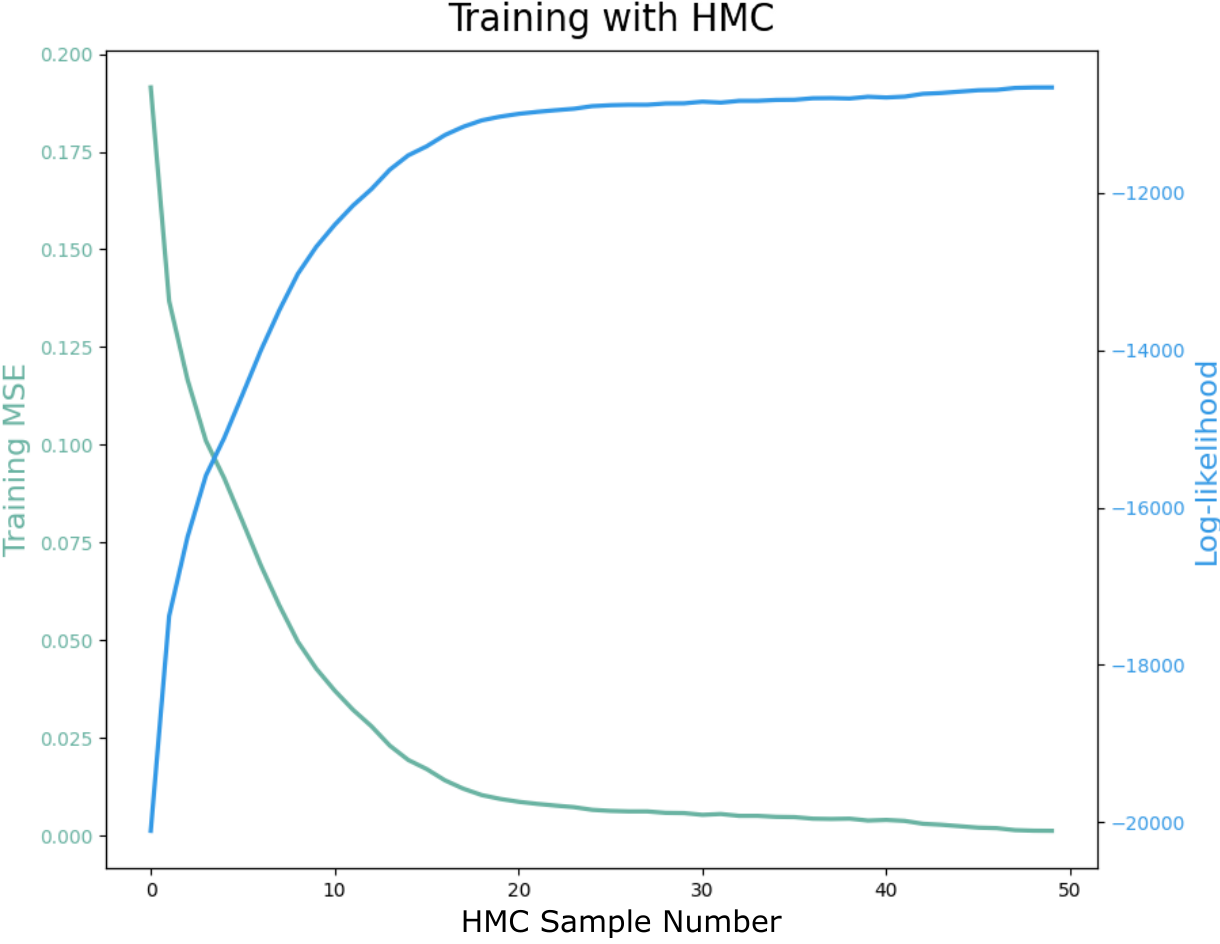}
\caption{Folded Cascode OTA Power Modeling}
\label{fig:hmc_res}
\end{figure}

\textbf{Schematic-Level Sizing Automation:} We tested our algorithm, BNN-BO, and other baseline algorithms on two circuits: two-stage folded cascode ota and strong-arm latch comparator. All transistors in both designs are parameterized for optimization. The Folded Cascode OTA has 20 independent design variables, and the Strong-Arm Latch Comparator has 13 independent design variables after respecting the symmetry constraints. The parameterized device sizes include: transistor lengths \& widths, capacitor values, and multipliers (integer valued). 

The schematic-level constrained sizing problem for Folded Cascode OTA is defined as follows:
\vspace*{0mm}
\small
\begin{equation}\label{specifications_folded}
\resizebox{0.92\columnwidth}{!}{%
$\begin{array}{l}
\text{minimize Power} \\
\text { s.t. } \text{\enspace DC Gain}>60 \mathrm{\enspace dB} \qquad {\text{Settling Time}<\mathrm{30} \mathrm{\enspace ns}} \\
    \qquad \begin{array}{l l}
    \text{CMRR}>80 \mathrm{\enspace dB}& {\text{Saturation Margin}>\mathrm{50} \mathrm{\enspace mV}} \\
    \text{PSRR}>80 \mathrm{\enspace dB} & \text{Unity Gain Freq.}>30 \mathrm{\enspace MHz}\\
    \text{Out. Swing}>2.4 \mathrm{\enspace V} & \text{Out. Noise}<\mathrm{30} \mathrm{\enspace mV_{rms}}  \\
    \text{Static error}<0.1 &\text{Phase Margin}>60 \mathrm{\enspace deg.}\\
    \end{array}
\end{array}$%
}
\vspace*{0mm}
\end{equation}
\normalsize

The schematic-level constrained sizing problem for Strong-Arm Latch Comparator is defined as follows:
\small
\begin{equation}
\label{specifications:SA}
\begin{array}{l}{\text { minimize } \text{Power}} \\
{\text { s.t. } \text{\enspace Set Delay}<10 \mathrm{\enspace ns}} \\
    {\qquad \begin{array}{l}{\text{Reset Delay}<6.5 \mathrm{\enspace ns}} \\
    {\text{Input-referred Noise}< \mathrm{50} \mathrm{\enspace \mu Vrms}} \\
    {\text{Differential Reset Voltage}< \mathrm{1} \mathrm{\enspace \mu V}} \\
    {\text{Differential Set Voltage}> \mathrm{1.195} \mathrm{\enspace V}} \\
    {\text{Positive-Integration Node Reset Voltage}< \mathrm{60} \mathrm{\enspace \mu V}} \\
    {\text{Negative-Integration Node Reset Voltage}< \mathrm{60} \mathrm{\enspace \mu V}} \\
    {\text{Positive-Output Node Reset Voltage}< \mathrm{0.35} \mathrm{\enspace \mu V}} \\
    {\text{Negative-Output Node Reset Voltage}< \mathrm{0.35} \mathrm{\enspace \mu V.}} \\   
    \end{array}}
\end{array}
\vspace*{0mm}
\end{equation}
\normalsize 

\begin{figure}[t]
\centering
\includegraphics[scale=0.6]{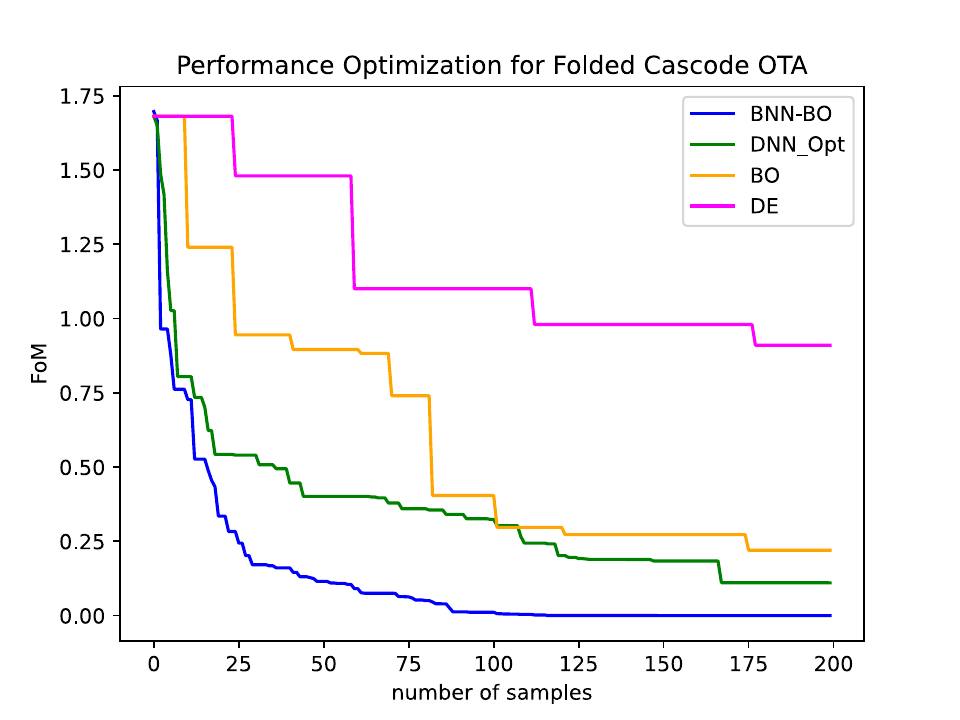}
\caption{Folded Cascode Optimization}
\label{fig:folded_fom}
\end{figure}

\begin{table*}[t]
\begin{centering}
    \caption{Schematic-Level Sizing Optimization Statistics}
    \label{table:schematic-combined-results}
\end{centering}
      
\begin{center}
\resizebox{1.8\columnwidth}{!}{%
\begin{tabular}{|l|cccc||cccc|}
    \hline
    Circuit Name & \multicolumn{4}{|c||}{Folded Cascode OTA}&\multicolumn{4}{|c|}{Strong-Arm Latch Comparator}\\
    \hline
    Algorithm & DE & BO & DNN-Opt &\textbf{BNN-BO}& DE & BO & DNN-Opt &\textbf{BNN-BO} \\
    \hline
    success rate & 10/10 & 7/10 & 10/10 & \textbf{10/10}& 7/10 & 1/10 & 9/10 & \textbf{10/10}\\
    \hline
    \# of simulations & 3200 & 340 & 151 &  \textbf{82}& 2800 & $>$500 & 154 &  \textbf{68}\\
    \hline
    Min power (m$W$) & 0.75 & 0.88 & 0.64  & \textbf{0.60}& 3.02 & 3.67 & \textbf{2.45}  & 2.5\\
    \hline
    Max power (m$W$) & 1.53 & 1.43 & 0.8 & \textbf{0.75}& 4.1 & 3.67 & 2.66 & \textbf{2.55}\\
    \hline
    Mean pow. (m$W$) & 1.14 & 1.19 & 0.72  & \textbf{0.69}& 3.44 & 3.67 & 2.54  & \textbf{2.52}\\
    \hline
    Modeling time (h) &NA & 30 &\textbf{0.6} & 1.5 & NA & 17 &\textbf{0.3} & 0.7\\
   \hline
   Simulation time (h) &54 &2.7 &2.7 &2.7 &72 &3.6 &3.6 &3.6  \\
    \hline
   Total runtime (h) &54 &32.7 &\textbf{3.3} &4.2& 72 & 20.6 &\textbf{3.9} & 4.3 \\
    \hline
\end{tabular}%
}
\end{center}
\end{table*}

We show the accuracy of the BNN modeling by demonstrating the training metrics. Training Mean Squared Error~(MSE) and the logarithmic likelihood of the fitted model are given in Figure~\ref{fig:hmc_res}. Collecting new HMC samples from the posterior increases the likelihood and reduces the training error. We observed very similar training schemes for all other circuits and performance metrics.

We repeat all experiments 10 times to account for the randomization involved in tested algorithms. The statistical results of our tests are shown in Table~\ref{table:schematic-combined-results}. Testing on both circuits suggests that BNN-BO can achieve feasible solutions in all runs, and it uses the smallest number of simulations to achieve this. Compared to Differential Evolution (DE), BNN-BO can find feasible solutions using up to 40x less number of simulations. Compared to the closest baseline algorithm, DNN-Opt, BNN-BO reduces the simulation time for finding similar results by up to $55\%$, proving its high efficiency. It is also demonstrated in Table~~\ref{table:schematic-combined-results} that, on average, the final design proposed by BNN-BO draws up to $40\%$ less power. The only disadvantage of BNN-BO to DNN-Opt is the modeling time as DNN-Opt maintains a single DNN model to approximate all performance metrics. Note that all reported times consider the full simulation budget~(500 new samples). Therefore, although it takes longer time for BNN-BO to do a single iteration, the required real time for BNN-BO to find a feasible solution is still smaller than other approaches.
\begin{figure}[t]
\centering
\includegraphics[scale=0.6]{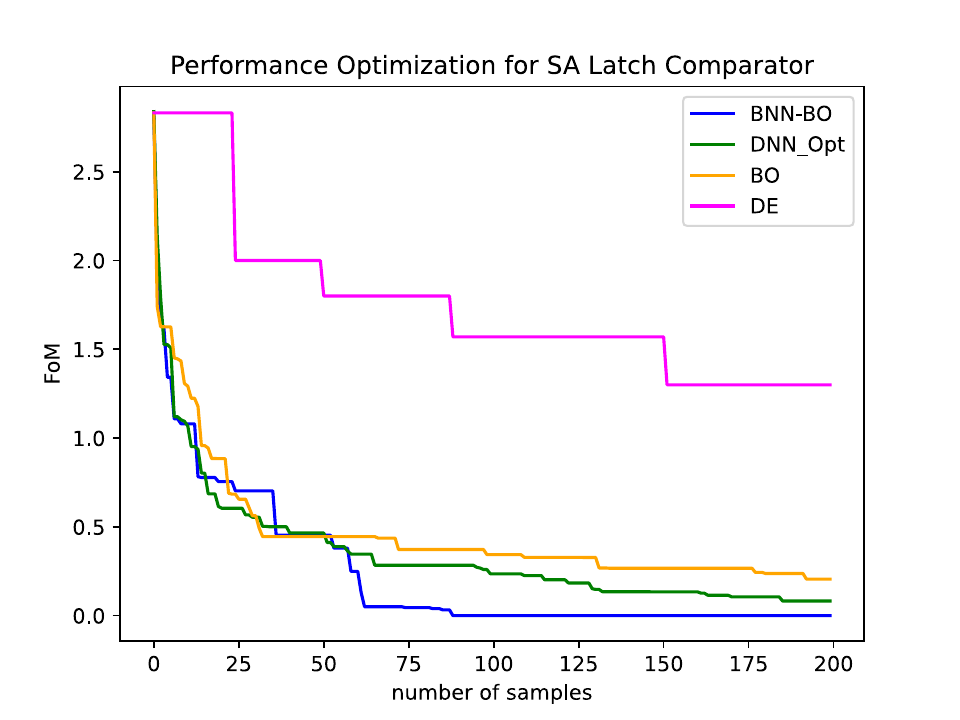}
\caption{Strong-Arm Latch Comparator Optimization}
\label{fig:sa_fom}
\end{figure}

In addition to experiment statistics, we further include the FoM convergence curves of both tests in Figure~\ref{fig:folded_fom} and Figure~\ref{fig:sa_fom}. The y-axis in the graphs represents the total constraint violation; therefore, FoM=0 represents a feasible solution. We observe that, compared to DNN-Opt, BNN-BO has $65\%$ and $33\%$ smaller area under the curve for Folded Cascode OTA and SA Latch Comparator, respectively.


\vspace{2mm}
\textbf{Layout-Aware Design Automation:}
In order to demonstrate the importance of layout effects on the final performance, we perform experiments on a Miller OTA circuit designed in 40nm technology (Fig.~\ref{fig:all_exp_cirs}). The optimization problem has 17 independent design variables and the optimization problem is defined as follows:

\begin{equation}\label{specifications_miller}
\resizebox{1\columnwidth}{!}{%
$\begin{array}{l}
\text { minimize } \text{Power} \\
\text { s.t. } \text{\enspace DC Gain}>45 \mathrm{\enspace dB} \qquad {\text{Settling Time}<\mathrm{100} \mathrm{\enspace ns}} \\
    \qquad \begin{array}{l l}
    \text{CMRR}>55 \mathrm{\enspace dB}& \text{Saturation Margins}>\mathrm{50} \mathrm{\enspace mV} \\
    \text{PSRR}>55 \mathrm{\enspace dB} & \text{Unity Gain BW.}>40 \mathrm{\enspace MHz}\\
    {\text{Out. Swing}>1 \mathrm{\enspace V}} & {\text{RMS Noise}<\mathrm{400} \mathrm{\enspace uV_{rms}}}  \\
    \text{Static error$<\%2$} &\text{Phase Margin}>60 \mathrm{\enspace deg.}\\
    \end{array}
\end{array}$%
}
\end{equation}

\begin{figure}[t]
\centering
\includegraphics[scale=0.39]{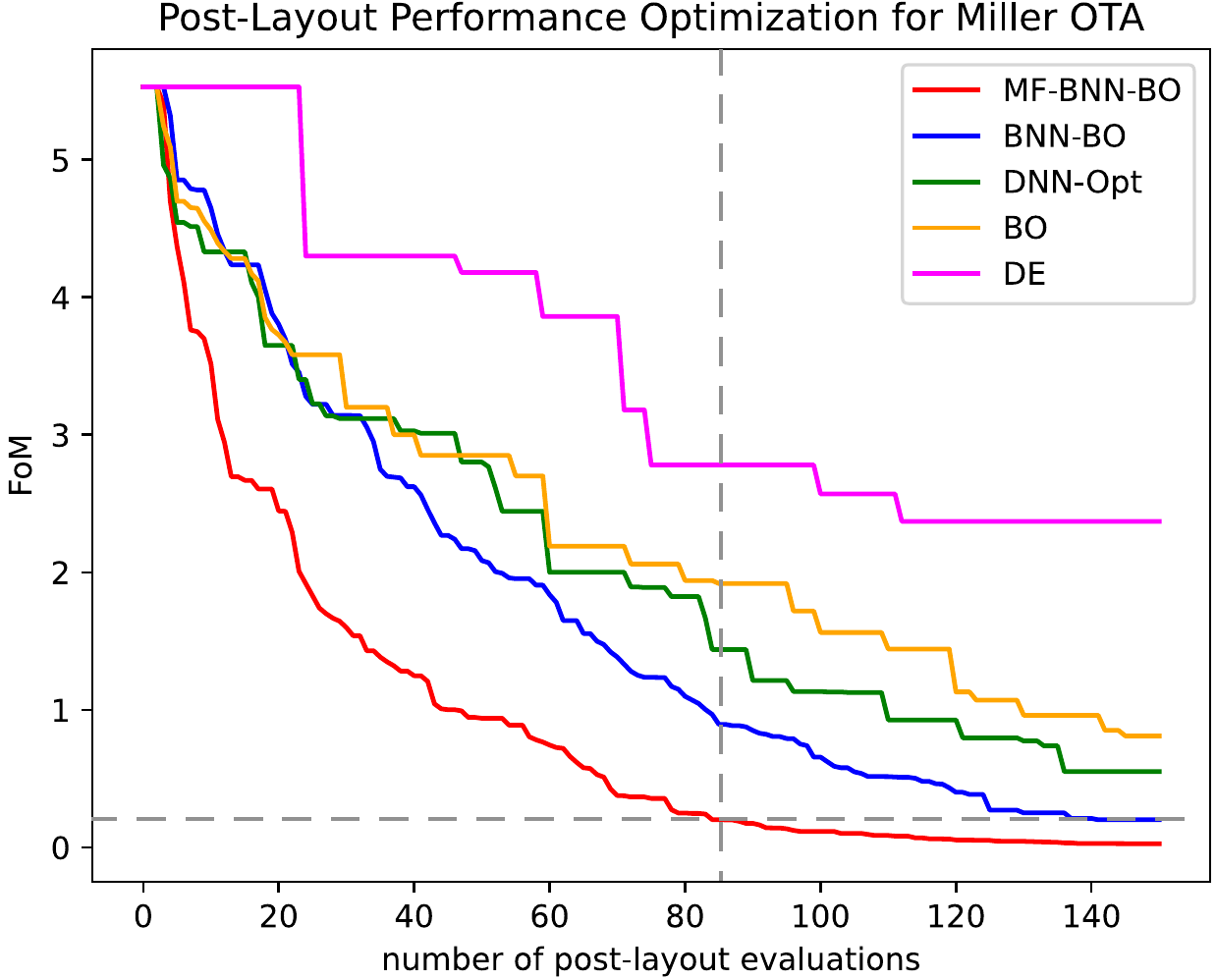}
\caption{Miller OTA Post-Layout Performance Optimization}
\label{fig:millerfom}
\end{figure}

Obtaining the post-layout performance of the Miller OTA is around 9 times more expensive than obtaining the schematic-level performance. Therefore this experiment is to prove the efficiency by utilizing multiple information sources. We initialize all algorithms with 50 high-fidelity random samples, and MF-BNN-BO has additional 50 samples from low-fidelity source~(schematic-level simulations). We demonstrate the FoM evolution for the rest of the optimization steps in Figure~\ref{fig:millerfom}. We observe that our Multi-Fidelity BNN algorithm provides even more efficiency compared to already efficient BNN-BO. Our analysis shows that the area under the curve is $45\%$ smaller for MF-BNN-BO compared to BNN-BO. Further, we observe that BNN-BO's average best solution after 150 high-fidelity iterations is surpassed by MF-BNN-BO only using 84 simulations. This also implies close to~$45\%$ improved efficiency due to utilizing correlations between the schematic-level evaluations and post-layout level evaluations. Note that there is an equal number of schematic-level simulations while running MF-BNN-BO that are not reflected in Figure~\ref{fig:millerfom}. Considering these simulations, the time efficiency is slightly reduced to around $38\%$. This efficiency figure serves as a lower-bound since we leave the improvements on fidelity selection as a future work.

\section{Conclusion} 
\label{sec:conclusion}
In this work, we presented Bayesian Neural Network-based solutions for schematic-level analog sizing automation and post-layout performance optimization. We targeted the scalability issue of the learning-based automation methods and provided a sample efficient optimization flow. We demonstrated the efficiency of the proposed approaches on academic benchmarks. Compared to the state-of-the-art, we improved the sizing automation efficiency by up to $45\%$. The Multi Fidelity BNN algorithm analysis proved that utilizing cheaper~(schematic-level) simulations reduces the need for expensive~(post-layout) simulations considerably, further boosting the efficiency.


{
\bibliographystyle{IEEEtran}
\bibliography{analog.bib}

\begin{thebibliography}{10}
\providecommand{\url}[1]{#1}
\csname url@samestyle\endcsname
\providecommand{\newblock}{\relax}
\providecommand{\bibinfo}[2]{#2}
\providecommand{\BIBentrySTDinterwordspacing}{\spaceskip=0pt\relax}
\providecommand{\BIBentryALTinterwordstretchfactor}{4}
\providecommand{\BIBentryALTinterwordspacing}{\spaceskip=\fontdimen2\font plus
\BIBentryALTinterwordstretchfactor\fontdimen3\font minus
  \fontdimen4\font\relax}
\providecommand{\BIBforeignlanguage}[2]{{%
\expandafter\ifx\csname l@#1\endcsname\relax
\typeout{** WARNING: IEEEtran.bst: No hyphenation pattern has been}%
\typeout{** loaded for the language `#1'. Using the pattern for}%
\typeout{** the default language instead.}%
\else
\language=\csname l@#1\endcsname
\fi
#2}}
\providecommand{\BIBdecl}{\relax}
\BIBdecl

\bibitem{10.1145/3569052.3578929}
A.~F. Budak, K.~Zhu, H.~Chen, S.~Poddar, L.~Zhao, Y.~Jia, and D.~Z. Pan,
  ``Joint optimization of sizing and layout for ams designs: Challenges and
  opportunities,'' in \emph{Proceedings of the 2023 International Symposium on
  Physical Design}, ser. ISPD '23.\hskip 1em plus 0.5em minus 0.4em\relax New
  York, NY, USA: Association for Computing Machinery, 2023, p. 84–92.

\bibitem{Vural2012AnalogCS}
A.~Vural \emph{et~al.}, ``Analog circuit sizing via swarm intelligence,''
  \emph{AEU - International Journal of Electronics and Communications}, 2012.

\bibitem{Liu:2013:ADA:2526263}
B.~Liu, G.~Gielen, and F.~V. Fernndez, \emph{Automated Design of Analog and
  High-frequency Circuits: A Computational Intelligence Approach}.\hskip 1em
  plus 0.5em minus 0.4em\relax Springer, 2013.

\bibitem{GASPAD}
B.~{Liu}, D.~{Zhao}, P.~{Reynaert}, and G.~G.~E. {Gielen}, ``Gaspad: A general
  and efficient mm-wave integrated circuit synthesis method based on surrogate
  model assisted evolutionary algorithm,'' Feb 2014.

\bibitem{Lyu:2018:MBO:3195970.3196078}
W.~Lyu, F.~Yang, C.~Yan, D.~Zhou, and X.~Zeng, ``Multi-objective bayesian
  optimization for analog/rf circuit synthesis,'' 2018.

\bibitem{9775179}
B.~He, S.~Zhang, Y.~Wang, T.~Gao, F.~Yang, C.~Yan, D.~Zhou, Z.~Bi, and X.~Zeng,
  ``A batched bayesian optimization approach for analog circuit synthesis via
  multi-fidelity modeling,'' \emph{IEEE Transactions on Computer-Aided Design
  of Integrated Circuits and Systems}, vol.~42, no.~2, pp. 347--359, 2023.

\bibitem{Rasmussen:2005:GPM:1162254}
C.~E. Rasmussen and C.~K.~I. Williams, \emph{Gaussian Processes for Machine
  Learning}.\hskip 1em plus 0.5em minus 0.4em\relax The MIT Press, 2005.

\bibitem{NIPS2012_05311655}
J.~Snoek, H.~Larochelle, and R.~P. Adams, ``Practical bayesian optimization of
  machine learning algorithms,'' vol.~25, 2012.

\bibitem{Analog_book2022_Budak}
A.~F. Budak, S.~Zhang, M.~Liu, W.~Shi, K.~Zhu, and D.~Z. Pan, ``Machine
  learning for analog circuit sizing,'' in \emph{Machine Learning for Analog
  Circuit Sizing}, H.~Ren and J.~Hu, Eds.\hskip 1em plus 0.5em minus
  0.4em\relax Springer, 2022, pp. 307--335.

\bibitem{MAGICAL_ICCAD19_Xu}
B.~{Xu}, K.~{Zhu}, M.~{Liu}, Y.~{Lin}, S.~{Li}, X.~{Tang}, N.~{Sun}, and D.~Z.
  {Pan}, ``{MAGICAL}: Toward fully automated analog {IC} layout leveraging
  human and machine intelligence,'' 2019.

\bibitem{Goan_2020}
E.~Goan and C.~Fookes, ``Bayesian neural networks: An introduction and
  survey,'' in \emph{Case Studies in Applied Bayesian Data Science}.\hskip 1em
  plus 0.5em minus 0.4em\relax Springer International Publishing, 2020, pp.
  45--87.

\bibitem{9434374}
A.~Budak, M.~Gandara, W.~Shi, D.~Pan, N.~Sun, and B.~Liu, ``An efficient analog
  circuit sizing method based on machine learning assisted global
  optimization,'' 2021.

\bibitem{Wang2020GCNRLCD}
H.~Wang, K.~Wang, J.~Yang, L.~Shen, N.~Sun, H.~Lee, and S.~Han, ``{GCN-RL}
  circuit designer: Transferable transistor sizing with graph neural networks
  and reinforcement learning,'' 2020.

\bibitem{Settaluri2020AutoCktDR}
K.~Settaluri, A.~Haj-Ali, Q.~Huang, K.~Hakhamaneshi, and B.~Nikoli{\'c},
  ``{AutoCkt:} deep reinforcement learning of analog circuit designs,'' 2020.

\bibitem{budak2021dnnopt}
A.~F. Budak, P.~Bhansali, B.~Liu, N.~Sun, D.~Z. Pan, and C.~V. Kashyap,
  ``{DNN-Opt} an {RL} inspired optimization for analog circuit sizing using
  deep neural networks,'' 2021.

\bibitem{9643445}
J.~Liu, S.~Su, M.~Madhusudan, M.~Hassanpourghadi, S.~Saunders, Q.~Zhang,
  R.~Rasul, Y.~Li, J.~Hu, A.~K. Sharma, S.~S. Sapatnekar, R.~Harjani, A.~Levi,
  S.~Gupta, and M.~S.-W. Chen, ``From specification to silicon: Towards
  analog/mixed-signal design automation using surrogate nn models with transfer
  learning,'' in \emph{2021 IEEE/ACM International Conference On Computer Aided
  Design (ICCAD)}, 2021, pp. 1--9.

\bibitem{8942174}
Z.~Gao, J.~Tao, F.~Yang, Y.~Su, D.~Zhou, and X.~Zeng, ``Efficient performance
  trade-off modeling for analog circuit based on bayesian neural network,'' in
  \emph{2019 IEEE/ACM International Conference on Computer-Aided Design
  (ICCAD)}, 2019, pp. 1--8.

\bibitem{LOURENCO2016316}
N.~Lourenço, R.~Martins, A.~Canelas, R.~Póvoa, and N.~Horta, ``Aida:
  Layout-aware analog circuit-level sizing with in-loop layout generation,''
  \emph{Integration}, vol.~55, pp. 316--329, 2016.

\bibitem{Analog_ICCAD19_Hakhamaneshi}
K.~{Hakhamaneshi}, N.~{Werblun}, P.~{Abbeel}, and V.~{Stojanović}, ``{BagNet}:
  Berkeley analog generator with layout optimizer boosted with deep neural
  networks,'' 2019.

\bibitem{Analog_DATE21_Liu}
M.~Liu, W.~J. Turner, G.~F. Kokai, B.~Khailany, D.~Z. Pan, and H.~Ren,
  ``Parasitic-aware analog circuit sizing with graph neural networks and
  bayesian optimization,'' 2021.

\bibitem{1206.1901}
R.~M. Neal, ``{MCMC} using {Hamiltonian} dynamics,'' \emph{Handbook of Markov
  Chain Monte Carlo}, vol.~54, pp. 113--162, 2010.

\bibitem{ErikssonP21}
D.~Eriksson and M.~Poloczek, ``Scalable constrained bayesian optimization,'' in
  \emph{The 24th International Conference on Artificial Intelligence and
  Statistics, AISTATS 2021, April 13-15, 2021, Virtual Event}, ser. Proceedings
  of Machine Learning Research, A.~B. 0001 and K.~Fukumizu, Eds., vol.
  130.\hskip 1em plus 0.5em minus 0.4em\relax PMLR, 2021.

\bibitem{NIPS2016_a96d3afe}
J.~T. Springenberg, A.~Klein, S.~Falkner, and F.~Hutter, ``Bayesian
  optimization with robust bayesian neural networks,'' in \emph{Advances in
  Neural Information Processing Systems}, D.~Lee, M.~Sugiyama, U.~Luxburg,
  I.~Guyon, and R.~Garnett, Eds., vol.~29.\hskip 1em plus 0.5em minus
  0.4em\relax Curran Associates, Inc., 2016.

\bibitem{NIPS2016_605ff764}
K.~Kandasamy, G.~Dasarathy, J.~B. Oliva, J.~Schneider, and B.~Poczos,
  ``Gaussian process bandit optimisation with multi-fidelity evaluations,'' in
  \emph{Advances in Neural Information Processing Systems}, D.~Lee,
  M.~Sugiyama, U.~Luxburg, I.~Guyon, and R.~Garnett, Eds., vol.~29.\hskip 1em
  plus 0.5em minus 0.4em\relax Curran Associates, Inc., 2016.

\bibitem{NEURIPS2019_9015}
A.~Paszke, S.~Gross, F.~Massa, A.~Lerer, J.~Bradbury, G.~Chanan, T.~Killeen,
  Z.~Lin, N.~Gimelshein, L.~Antiga, A.~Desmaison, A.~Kopf, E.~Yang, Z.~DeVito,
  M.~Raison, A.~Tejani, S.~Chilamkurthy, B.~Steiner, L.~Fang, J.~Bai, and
  S.~Chintala, ``Pytorch: An imperative style, high-performance deep learning
  library,'' in \emph{Advances in Neural Information Processing Systems
  32}.\hskip 1em plus 0.5em minus 0.4em\relax Curran Associates, Inc., 2019,
  pp. 8024--8035.

\bibitem{balandat2020botorch}
M.~Balandat, B.~Karrer, D.~R. Jiang, S.~Daulton, B.~Letham, A.~G. Wilson, and
  E.~Bakshy, ``{BoTorch: A Framework for Efficient Monte-Carlo Bayesian
  Optimization},'' in \emph{Advances in Neural Information Processing Systems
  33}, 2020.

\bibitem{cobb2020scaling}
A.~D. Cobb and B.~Jalaian, ``Scaling hamiltonian monte carlo inference for
  bayesian neural networks with symmetric splitting,'' \emph{Uncertainty in
  Artificial Intelligence}, 2021.

\end{thebibliography}
}



\end{document}